\documentclass[runningheads]{llncs}

\usepackage{graphicx}
\usepackage{amssymb}
\usepackage{amsmath}
\usepackage{color}
\usepackage{babel}
\usepackage{algorithm}
\usepackage{algorithmic}

\usepackage{todonotes}	% uncomment to show all the notes 

\usepackage{subcaption} 
\begin{document}
\title{End-Effect Exploration Drive for Effective Motor Learning}
\author{Emmanuel Daucé\inst{1,2}\orcidID{0000-0001-6596-8168}}
\authorrunning{E. Daucé}
\institute{Institut de Neurosciences de la Timone, CNRS/Aix-Marseille Univ, France
	\and
Ecole Centrale de Marseille, France\\ 	
\email{emmanuel.dauce@univ-amu.fr}}
\maketitle              % typeset the header of the contribution
\begin{abstract}
%We develop a learning framework that implements the inversion of a target distribution
%of effects to train a controller. When the target distribution is uniform, this
%setup provides an efficient exploration policy. The main idea is to use and maintain a simplified effect model together with the online update of the policy.
%Combined with an extrinsic reward, it is shown to provide a faster 
%training than traditional off-policy techniques. 
%The framework is mostly devoted to the case of motor learning, when
%the number of degrees of freedom is large and the final effect space
%is small and bounded. 
Stemming on the idea that a key objective in reinforcement learning is to invert a target distribution of effects, end-effect drives are proposed as an effective way to implement goal-directed motor learning, in the absence of an explicit forward model.  An end-effect model relies on a simple statistical recording of the effect of the current policy, here used  as a substitute for the more resource-demanding forward models. When combined with a reward structure, it forms the core of a lightweight variational free energy minimization setup. The main difficulty lies in the maintenance of this simplified effect model together with the online update of the policy. When the prior target distribution is uniform, it provides a ways to learn an efficient exploration policy, consistently with the intrinsic curiosity principles. When combined with an extrinsic reward, our approach is finally shown to provide a faster training than traditional off-policy techniques.  
\end{abstract}
\section{Introduction}

Recent developments in artificial intelligence have produced important qualitative leaps in the field of pattern recognition, %(identification of objects, faces, speech recognition, ...) 
in video games and assisted driving.
%{\color{magenta} [REF]}. 
%They have also enabled the emergence of generative models for images, video and language.
% {\color{magenta} [REF]}. 
%The key element that explains the success of these algorithms is the extraction of statistical invariants from data, via auto-encoding mechanisms, from data sets of arbitrary dimensions. The codes are constructed during training using the gradient descent technique. After learning, they provide synthetic descriptors of the situations encountered, which are necessary and sufficient for the production of appropriate responses.  Once learned, they thus constitute very powerful generative models of the input data. Nevertheless, the learning of these invariants currently requires a considerable amount of data, real or simulated, which makes the learning algorithms dependent on the available databases, and gives the data collection (or simulation) process a disproportionate place. Furthermore, while the capabilities of these algorithms approaches (and sometimes exceeds) those of human subjects for specific tasks, they never achieve the universal skills and general adaptability of the human brain. 
However, a number of tasks considered as simple, %shared across most living species, 
are struggling to find a convincing artificial implementation... This is the field of action selection and motor control. For instance, the fine manipulation of objects, as well as movement in natural environments, and their combination through real time motor control, remain major scientific challenges at present.
Compared to the case of video games, reinforcement learning, for instance, remains rather limited in the field of robotic and motor control. The huge  improvements ``from-scratch'' obtained in virtual environments are difficult to transfer to real robotics, where millions of plays can not be engaged under risk of breakage, and simulators are expensive to develop.
%On the other hand, the primary function of the brain is to enable movement, to orientate the body and try to use the environment at its own advantage. 
The brain capability to develop motor skills in a very wide range of areas  has thus no equivalent in the field of artificial learning.   
%The control of an articulated body is particularly complex, and if we consider humans, it takes about a year to learn bipedal walking. 
%The brain's ability to learn new abilities and skills is particularly striking, but it is not specific to humans. The brain in general adapts to its environment. Animals modify their behaviour flexibly according to their own history. 

%The nervous system is characterized by a high degree of noise and variability: synapses are stochastic, neurons are fatiguable, synaptic connections are plastic. Neurons taken at the individual level are extremely unreliable as a unit of information processing (to the same extent as would be biologically possible). There is an "excess" of noise in the nervous system, which leads to (i) an ability to "produce" noise/information; (ii) an ability to "sample" the response space; (iii) a principle of scan/exploration/collection (foraging) of the sensory/motor space (one does not immediately return to where one has already explored).

%Physiology
%\begin{itemize}
%	\item >80\% reciprocal connections in the cortex (Van Essen et al, 1992)
%	\item Intrinsic Activity (Fox et al, 2006)
%	\item Stochastic Behaviour (Shadlen, Newsome, 1998)
%	\item Plasticity of Hebb (Bi and Poo, 1998)	
%\end{itemize}
%Motor control
%\begin{itemize}
%	
%\end{itemize}

In short, learning a motor command, or a motor skill, happens to become difficult when considering a moderately complex set of effectors, like a multi-joint arm for instance, operating in the physical world. One aspect of the problem is that the set of circuits that process the sensory signals and produce a motor response is composed of digital units, the neurons, operating at fast pace, and adapting rapidly, while, on the other side, the operation space is made of many joints, rigid elements and muscles covering a wide, continuous domain of responses with many degrees of freedom and much longer response times. 

The reinforcement learning framework \cite{sutton2018reinforcement} provides a very generic setup to address the question of learning, both from a machine learning and the brain modelling perspectives. It contains many of the interesting constraints that an agent is facing in order to learn a motor act.
The theory, however, is constructed around digital (discrete) control principles.
The aim of a digital controller is to establish a one to one correspondence between stimuli and actions, by
matching the input with a template action. 
Given a set of pre-defined actions, an agent is expected to pick the one that matches the most the input.  
In order to learn how to act, the agent is guided by a reward signal, that is much cheaper to extract than an exact set point.
%much easier to train from data. 
%But : the rewards can be sparse.
Then the choice of the action is monitored by a scalar quantity, the utility, that is the long term sum of rewards \cite{sutton2018reinforcement}.
A Reinforcement Learning agent is agnostic about what the world is. It is just acting so as to maximize the utility.
Supported by the solid principles of dynamic programming, this agent is expected to end-up in an optimal motor behavior with regards to the reward constraints provided by the environment. 
%in principle easier to train than an analog one, but
A hidden difficulty however lies in finding a good training dataset for the agent. A classical problem in that case is the lack of sampling efficacy due to the sparsity of the rewards, or to the closed-loop structure of the control task, where the examples encountered are statistically dependent on the controller parameters, 
providing a risk of a self referential loop and local optimum.

%Formally speaking, the set of circuits, neurons that process the sensory signals and produce a motor response
%is the \emph{control space} and the body, muscles and environment is the \emph{operation space}. 
%The circuits and neurons are digital.
%The body and environment is a continuous physical system 

% Learning motor skills : many caveats. Temporal credit assignment.  

%There are generally two ways to design controllers.  
The counterpart to digital control is analog control, corresponding to the ``classic'' way to design a controller in the majority of real-world scenarios. In the analog control case, both the controller and the environment are dynamical systems. 
%More precisely : a controller is organised around its sensors / actuators with a measure and 
%an action. 
%The controller acts on the environment (change according to internal states). 
%The environment acts on the controller (that changes according to its internal state -- memory)
The control of a dynamic system generally relies on a \emph{model} \cite{miall1996forward}. 
The controller is capable to mimic the
behavior of the environment, and know the effect of its actions on the environment. 
When a certain objective is given, the controller can 
act so as to reach the objective through \emph{model inversion}, though, in most cases of interest (like the control of an articulated body), the models are not invertible and no single definite control can be established from a given objective \cite{jordan1992forward} without setting up additional and task-specific regularization constraints. 
This kind of controller needs a forward model, that is generally given, containing a large engineering knowledge about the agent's effector structure and its environment. 
A problem arises when the learning of a motor command is considered. Such controllers generally lack in adaptivity, and motor adaptation is generally hard to train when the environmental 
 conditions change. 
 
There is thus an apparent trade-off between, on one side, maintaining a model, that may include the many mechanical, environmental and sensory data interactions, and, on the other side,  being guided by a simplistic reward, concentrating all the complexity of the external world constraints in a single scalar.
The main difference between both approaches appears to be the presence (or the absence) of a definite model of the external world. Knowing the effect of our own actions in the world provides ways to anticipate and do planning to reach an objective, at the cost of maintaining and updating a model of the world \cite{kurutach2018model}. 
%{\color{magenta} [PROBLEMATIQUE : Model free / model based. Le "model-free adopté par les "main players". Le model-based est l'approche adoptée par l'école de Friston il me semble]}
At present time, the trade-off between model-free and model-based control has provoked many debates in the reinforcement learning community, with a preference toward model-free approaches for they are cheaper to maintain and easier to control, leaving unsolved the problem of the sampling efficacy and causing very long training sessions in moderately complex environments.
 
Our argument here is that the sampling efficacy is bound to the problem of training a model, and one can not expect to have an efficient sampling without a minimal model of the \emph{effect} of action. This model does not need to be perfectly accurate, but it needs to be good enough to allow the agent to efficiently sample its environment in order to grab all the disposable information in relation to the task at hand. We assert here that a simple \emph{effect model} is enough to provide all the needed variability in probing the effect of an action or a policy.

%	\item Spatial and temporal compositionality of motor tasks (Diedrichsen \& Kornysheva, 2015, Kadmon Harpaz et al, 2019)

%{\color{magenta} Our approach is different from the `curious'', and ''novelty seeking'' agents.}
%{\color{blue}
%In order to circumvent this problem, a body of work has developed curiosity and novelty-seeking in agents, independently from the utility of actions.  [REFS]
%Novelty seeking relies on building a model of the environment to which the current data is compared. A data showing inconsistencies with the model is considered as "interesting" and should attract the attention of the agent, independently of its extrinsic value.  }

%{\color{magenta} Plan?}

\section{Method}

\subsection{A probabilistic view to motor supervision}

The probabilistic view to learning cause-effect relationships is at the core of many recent developments in machine learning, with a body of optimization techniques known as “variational inference” implementing model training from data \cite{kingma2013auto}. 
%One key idea is here to take inspiration from some theories developed in machine learning to reconsider the brain at the light of probabilistic inference and optimization.
%A first and not  yet fully unveiled example is how the brain learns the consequence of its own action though experiencing interactions with the world [2]. A second example in vision is how the eye selects its next saccade in order to deepen the understanding of its visual environment [3].
%Taking motor control learning and action selection as a main guideline, our general objective is to decipher the respective contributions of information seeking \cite{mohamed2015variational}, learning improvement \cite{schmidhuber1991curious} and utility maximization \cite{sutton2018reinforcement} in the selection of actions in an ecological setting, with both perspectives in modelling brain function and developing innovative brain-inspired machine learning techniques for control and robotics.
We assume for simplicity that the environment is not hidden to the agent, i.e. the environment is fully observable. We also assume a discrete updating of states and actions, like in classic reinforcement learning.
Then, if $s$ is the state of the environment (or a context), and $a$ an action performed by the agent, consider $e$ as the \emph{effect} of the action performed by the agent in that particular state. 

The effect may reflect, to some point, the result,  or the outcome, of the action. Modelling an effect thus supposes that an action should come to an end, a final point, from which it is possible to evaluate or record a result. Consistently with a bunch of neurological observations \cite{graziano2002complex}, a simple end-effector open-loop control is here assumed to take place, with a compositional motor command \cite{kadmon2019movement} driving a multi-joint effector toward a fixed point, without feedback during the motor realization. 

%{\color{magenta} pour une plus grande généralité, on mixe état et contexte d'une part, effet et état marginal de l'autre. L'action est par principe une action composée.} 

The effect can be a short-term effect, like reading a new state from the environment. It can also be a long term effect, like winning or loosing a game, or reaching an objective $s^*$ in the future. Because there is a lot of uncertainty on the effect of an action, it is modeled as a probability distribution $p(E|s,a)$.
When the effect is not dependent on a context, it can be noted more simply $p(E|a)$.
Given a certain policy $\pi(a|s)$, one can also consider the average distribution of effects obtained when applying that specific policy, namely 
\begin{align}\label{eq:effect-model}
p(E|s) = \mathbb{E}_{a\sim \pi(A|s)} p(E|s,a)
\end{align}  
This marginal distribution is said the effect distribution. By construction, it is dependent on a particular policy (possibly stochastic) over states and actions, and may, for instance, represent the invariant measure of an MDP under that given policy. In our case however, we mostly consider the open-loop control case. The policy is defined over the elements of a compositional action, that is choosing the components of an action. The end-effect of such a compositional action is the terminal state attained at the end of the action, without reading the intermediate states.  
   
%{\color{magenta} Cette distribution est appelée ici la distribution des effets. Il est important de souligner la dépendance à la politique menée. Analogie : la mesure invariante d'une chaîne de Markov. Mettre des  refs en particulier : state marginal matching [Lee et al].}

In goal-directed control, if $e$ is an expected effect, an \emph{inverse control policy}, whose role is to maximize the chance to reach the effect, can be defined using Bayes rule as:
\begin{align}\label{eq:inv-policy}
\pi(a|s,e) = \frac{p(e|s,a)\pi(a|s)}{p(e|s)}
\end{align}
That is the inversion of the model in a probabilistic setup \cite{bays2007computational}. Here the marginal effect distribution plays the role of a set point, that fixates the distribution of states toward which the action should head for.
%

%{\color{magenta} La distributon marginale joue un rôle très important. C'est elle qui dicte l'objectif.}

Assume now $p^*(e|s)$ be a target distribution of effects. This distribution is distinct from  $p(e|s)$ that is the distribution of effects under the current policy. It is assumed to be realizable from an (unknown) target policy $\pi^*(a|s)$, that can be decomposed into:
\begin{align}
\pi^*(a|s) &= \mathbb{E}_{e\sim p(E|s,a)} \pi^*(a|s,e)\frac{p^*(e|s)}{p(e|s,a)}
\end{align}

The right side of the equation provides an estimation of the optimal policy based on a sample $e$ of the effect of the action. Unfortunately, the optimal inverse control policy $\pi^*(a|s,e)$ is unknown. A shortcut is to approximate it with the current inverse control policy $\pi(a|s,e)$.
In that case, it happens from equation (\ref{eq:inv-policy}) that the formula simplifies to :
\begin{align}\label{eq:bayesian-update}
\pi^*(a|s) &\simeq \mathbb{E}_{e\sim p(E|s,a)} \pi(a|s,e)\frac{p^*(e|s)}{p(e|s,a)}\\
&= \pi(a|s) \mathbb{E}_{e\sim p(E|s,a)} \frac{p^*(e|s)}{p(e|s)}
\end{align}
This formula shows that a correction term can be applied to the current policy without owning an explicit forward model, but rather through reading the average effect of the policy. This forms the basis of a supervised approach to motor learning, allowing to update a policy so as to reach a target marginal distribution. 

%Consider now that the current policy $\pi$ is parameterized by a set of parameters noted $\theta$. For simplicity, we assume in the following that $\theta=(Q,\beta)$ i.e. $\forall s,a$,
%\begin{align}
%\pi(a|s) = \frac{\exp \beta Q(s,a)}{\sum_{a'\in \mathcal{A}} \exp \beta Q(s,a')}
%\end{align}
%that is the softmax policy parameterized by an action-value function $Q$ and an inverse temperature $\beta$.
%
%Then, noting $Z= \sum_{a'\in \mathcal{A}} \exp \beta Q(s,a')$, 
%\begin{align}
%\log \pi(a|s) = \beta Q(s,a) - \log Z
%\end{align}
%
%From equation (\ref{eq:bayesian-update}), it comes:
%\begin{align}
% Q(s,a) - Q^*(s,a) &\simeq %\frac{1}{\beta} \left(\log \mathbb{E}_{e\sim p(E|s,a)} \frac{p(e|s)}{p^*(e|s)} + \log Z - \log Z^*\right)\\
% %&\leq 
% \frac{1}{\beta}  \left(\mathbb{E}_{e\sim p(E|s,a)} [\log p(e|s) - \log p^*(e|s)] + \log Z - \log Z^*\right) \label{eq:TD-err}
%\end{align}
%making it possible to define a (biased) estimator of the \emph{error} on $Q(s,a)$ from the sampling of its effect.
%

For instance, in a dicrete setup, assuming there exists a $Z(s) \in \mathbb{R}$ such that $\log \pi(a|s) = \beta Q(s,a) + Z(s)$ (softmax policy) makes it possible to update $Q(s,a)$ with the last effect sample $e$ like:
\begin{align}\label{eq:Q-update}
Q(s,a) \leftarrow & Q(s,a) - \frac{\alpha}{\beta} (\log p(e|s) - \log p^*(e|s))
\end{align}  
This update renders the current policy closer to the optimal one. A side effect of this update is that it also changes the effect model that includes the contribution of the policy (see eq. (\ref{eq:effect-model})). Repeating the operation with a small $\alpha$ and different samples of $a$ and $e$ should, on average, reduce the divergence between $\pi$ and $\pi^*$.
%\subsection*{Link with reinforcement learning}
Equation (\ref{eq:Q-update}) also provides an interesting identity when considering the classical (reward-based) TD-error, with $\frac{1}{\beta} (\log p(e|s) - \log p^*(e|s))$ taking the role of the TD error, i.e. being identified with $Q(s,a)- \tilde{R}(e)$ (with $\tilde{R}(e)$ a putative sum of rewards up to $e$),
making it possible, for instance, to set up an intrinsic reward implementing a policy that realizes a known prior on the effects. 
This intrinsic reward is called here the ``End-Effect Drive''.

This supervised approach to policy relies on an effect model $p(e|s)$ that is less detailed than a forward model. Various kinds of approximate forward models can be found in goal-directed motor control literature, like dynamic goals \cite{kaelbling1993learning} and distal teacher \cite{jordan1992forward}, though generally learning to associate the current action with a distal effect \cite{mishra2017prediction,kurutach2018model}. In our case, the model knows nothing about the actions that are performed by the agent. Only the end-effects are recorded to build the model. This ``action-agnostic'' forward model is close to the concept of state-visit counter, as it is proposed in \cite{bellemare2016unifying}.

\subsection{A uniform exploration drive}

An important special case is when the objective is not to reach a certain effect, but rather to explore uniformly the range of all possible effects. In that case, the objective effect distribution $p^*$ is uniform over the effect space. This kind of supervision can be seen as a generalization of multiple-goal supervision \cite{haarnoja2017reinforcement} toward defining each possible effect as a goal.  The expected outcome of this uniform drive is to provide a uniform sampling over the effect space, i.e. implement a uniform exploration policy. This intrinsic reward is called here the ``End-Effect Exploration Drive'' (E3D in short). It is consistent with the pseudo-count bonus proposed in \cite{bellemare2016unifying}. A similar drive was also proposed in a recent draft as the ``state marginal matching'' drive \cite{lee2019efficient}.

%Propriétés d'un drive uniforme : un état très rarement visité apporte un reward très élevé. Inversement, état très fréquemment visité apporte un reward négatif. La distri uniforme est un point d'équilibre. Les rewards sont en principe équilibrés.

By construction, the E3D is positive when $e$ is rarely visited under the current policy ($p(e|s) < p^*(e)$), and negative the other way. It thus tries to promote rare and ``surprising'' effects, and lower the occurrence of habitual ``boring'' effects. It must be noticed that the promotion of rare effects tends to make them less rare, and the rejection of habitual effects tends to make them less habitual, up to an equilibrium where the log-probability ratio should be close to zero. 
Though the circular dependence between the policy update and the effect model update can provoke some convergence issues, and the 
 equilibrium may not be reached in the case of too fast fluctuations of both distributions during the training process. 
Some form of regularization is needed in most cases, and, most importantly, should be counterbalanced with some form of utility drive, in order to implement policy optimization through reward maximization.
This is the reason why a variational inference setup is particularly well suited in that case, with the distal uniform drive taking the role of a prior under a variational formulation.

\subsection{Link with variational inference}
A key intuition in Friston's  ``unified brain theory'' paper \cite{friston2010free} is interpreting the utility, as it is defined in economy and reinforcement learning, as a measure of the negative surprise (i.e the log probability over the sensory data distribution). Combined with a prior distribution in the control space, the action takes the role of a latent variable that is updated so as to reduce the prediction error with regards to the prior, much like in predictive coding. 

The unified approach proposed by Friston and colleagues is more generally consistent with the variational auto-encoder principles, in which a latent description of the data is constructed so as to implement a trade-off between the complexity of the description and the accuracy of the prediction. Variational reinforcement learning was recently proposed as a way to reconcile the discrete view to motor control with the continuous processing of the latent variable in variational auto-encoders \cite{levine2013guided,fox2015taming,haarnoja2017reinforcement}, with the motor command playing the role of a latent code for the reward data.  In our simplified writing, the utility maximization (or surprise minimization) rests on minimizing:
%\begin{equation}
%-\mathbb{E}_{a\sim \pi(a|s)} \log p(s|a) + \text{KL}(\pi(a|s)||\pi^*(a)) 
%\end{equation}
%with the left term being identified with the negative Q value (or action value) and the right term representing a regularization of the policy over the action space. Assuming that reading the effect provides a measure of the future rewards, it can be rewritten as:
\begin{equation}
-\mathbb{E}_{a\sim \pi(a|s); e\sim p(e|s,a)} \beta R(e) + \text{KL}(\pi(a|s)||\pi^*(a)) 
\end{equation}
with $R(e)$ here a measure of the sum of (extrinsic) rewards, up to $e$. Interestingly, the current policy $\pi(a|s)$ lies at the crossroad of a reference (maximum entropy) policy $\pi^*$ and reward maximization, with the softmax policy representing a compromise between both tendencies in the discrete case. 

Extending toward a uniform prior over the space of effects can be written when both considering the motor command and the effect as latent variables that may both explain the current observation, that writes:
\begin{equation}
-\mathbb{E}_{a, e\sim p(a,e|s))} \beta R(e) + \text{KL}(p(a,e|s))||p^*(a,e)) 
\end{equation}   

For the purpose of illustration, we propose here an additional simplification, that is assuming an \emph{independence} of both factors ($e$ and $a$) on causing the current data (Naïve Bayes assumption), dividing the KL term in two parts:
\begin{equation}
-\mathbb{E}_{a\sim \pi(a|s); e\sim p(e|s,a)} \beta R(e) + \text{KL}(\pi(a|s))||\pi^*(a)) + \text{KL}(p(e|s))||p^*(e))
\end{equation}   
This forms the baseline of our variational policy update setup. The optimization, that is done on $\pi(a|s)$, obeys in that case on a triple constraint, that is maximizing the reward through minimizing both the distance to a baseline policy and the distance of the effect to a reference (supposedly uniform) effect distribution.

%{\color{magenta} Ajouter l'expression de la free energy}

In a discrete setup, the uniform prior on the action is supposed implemented with the softmax decision rule. It is then sufficient to assume the following update for the action-value function.
After reading $e$, %the update of $Q$ should look like :
%\begin{align}
%Q(s,a) \leftarrow (1-\alpha) Q(s,a) + \alpha  (R(e) + \frac{1}{\beta} (\log p^*(e) - \log p(e|s)))
%\end{align}
%and 
the TD-error should be defined as:
\begin{align}\label{eq:var-RL-lambda}
\text{TD}(s,a,e) = \lambda(Q(s,a) - R(e)) + \frac{1}{\beta} (\log p(e|s) - \log p^*(e)))
\end{align}
With $\lambda$ a precision hyperparameter accounting for the different magnitudes of rewards, allowing to manipulate the balance between reward seeking and exploration-seeking drives.
Interestingly, the reward $R(e)$ has here the role of a regularizer with regards to the current Q-value. The sum of the future rewards can be estimated using the classical Bellman recurrence equation, i.e. $Q(s,a) \sim r(s,a) + Q(s', a')$, in which case the training procedure needs to maintain an estimate of a standard action-value function $Q_\text{ref}(s,a)$ to update the actual parameters of the policy $Q(s,a)$.

\section{Results}

% TODO : 1. Open loop (+ bras robot?) / 2. closed loop / 3. closed loop + amortized

%{\color{magenta} Présenter le cas épisodique / sans contexte / end-effector control. Forte différence avec la littérature classique de l'apprentissage par renforcement. La plupart des papiers considèrent la distributin distale des effets dans le cas récurrent (mesure invariante dela chaîne de Markov). OPEN-LOOP CONTROL. On a affaire à une séquence et pas  une chaîne de markov (plus précisément une chaîne "ouverte"). Pas de récurrence ici!!}

We present in simulation a pilot implementation of the principle presented in the previous section. The principal idea is to illustrate an important feature of biological motor control, that is the control of an effector showing many degrees of freedom, like e.g. an articulated limb with many joints. 
%In classical reinforcement learning, the action spaces are generally considered smaller than the sensory space. This is for instance the case when training a model on vintage Atari games, where the control is limited to a set of action keys \cite{mnih2013playing}.

Let $\mathcal{A}$ a control space accounting for a single degree fo freedom (here a discrete set of actions i.e. $\mathcal{A}=\{E,S,W,N\}$), each motor command owning $n$ degrees of freedom, i.e. $a_{1:n} = {a_1,...,a_n} \in \mathcal{A}^n $. The effect space is expected to be much smaller, like it is the case in end-effector control, where only the final set point of a movement in the peripheral space is considered as the result of the action. Each degree of freedom is supposed to be independent, i.e. the choice of $a_i$ does not depend on the choice of $a_j$, so that $\pi(a_{1:n}|s_0) = \pi(a_1|s_0) \times ... \times \pi(a_n|s_0)$. When a single context $s_0$ is considered, the policy writes simply $\pi(a_{1:n})$.
The size of the action space is thus combinatorially high, and one can not expect to enumerate every possible action in reasonable computational time. 
In contrast, the effect space is bounded, and the number of all final states can be enumerated. However, the environment is constructed in such a way that some final states are very unlikely to be reached under a uniform sampling of the action space.

%{\color{magenta} TODO : repréciser le lien avec "end-effector control" et boucle ouverte.}

The environment we consider is a grid world with only 18 states and two rooms, with a single transition allowing to pass from room A to room B (see figure 1). %\ref{fig:grid1}). 
Starting in the upper left corner of room A, the agent samples a trajectory $a_{1:7} \in \mathcal{A}$ from a policy $\pi$, that trajectory being composed of 7 elementary displacements. The agent does not have the capability to read the intermediate states it is passing through, it can only read the final state after the full trajectory is realized. 
In such an environment, a baseline uniform exploration does not provide a uniform distribution of the final states. In particular, when acting at random, the chance to end-up in the first room is significantly higher than the chance to end up in the second room.  
    
The agent starts from scratch, and has to build a policy $\pi(a)$ and an effect model $p(s_n)$, with $s_n$ the final state. 
There are two task at hand. A first task is a simple exploration task and the objective is to uniformly sample the effect space, which should imply a non-uniform sampling policy. A second task consists in reaching the lower-right corner, the state that shows the lowest probability with a uniform sampling. For that, a reward of 1 is given when the agent reaches the lower corner, and 0 otherwise.

\begin{figure}[t]\label{fig:grid1}
\centerline{
	\includegraphics[width = .4 \linewidth]{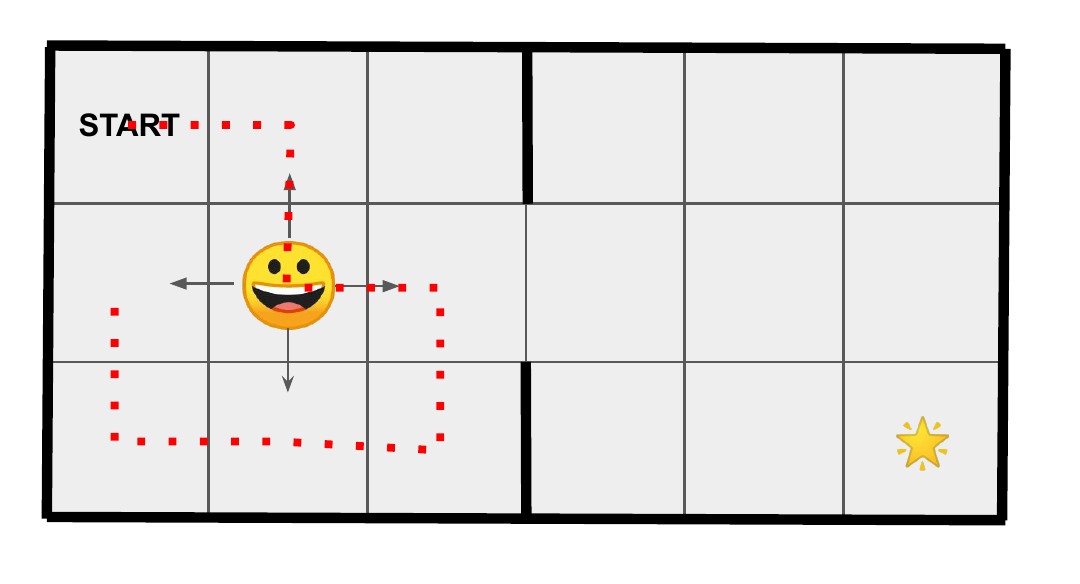} 
}
\caption{A simple two-rooms environment. Starting from the upper-left corner, the agent is asked to plan a full sequence made of 7 elementary actions $a_1,...,a_7$, each elementary action being in (E,S,W,N). The only read-out from the environment is the final state, and a reward, that is equal to 1 if the final state is the lower-right corner, and 0 otherwise.}
\end{figure}

The update procedure is made of a single loop (see algorithm \ref{algo:E3D}). The update is done online at each trial. Both the policy and the effect model are updated, with different training parameters. The general idea is to train the effect model a little more ``slowly'' than the policy, for the policy improvement to firmly take place before they are passed on the effect model. 

Stemming from a uniform policy, the effect of the E3D drive is to render the "rare" states more
attractive, for they are bound with a positive intrinsic reward, while the more commonly visited 
 states are bound with a negative reward, that reflects a form of "boredom". Marking a rare state as ``attractive'' tends to increase the number of visits, and finally lower the initial positive reward. In the case of a ``gradient'' in the likelihood of the final states, with a number of final visits inversely proportional to the distance to the initial state, the E3D drive favors a progressive ``expansion'' of the visiting territory, for each peripheral state attained will increase the probability to reach its further neighbors, up to the final limit of the state space. In small environment like the one proposed here, the limit is rapidly attained and a rapid alternation of visits is observed over the full state space. 

The final distribution of states is compared in figure 2 %\ref{fig:explore} 
in the case of a uniform policy and the E3D drive. In that specific setup, a strong bias in favor of the first room is observed, and a gradient of likelihood is observed from the initial state toward the lower right corner (figure 2A). %\ref{fig:explore}A). 
In contrast, a time consistent uniform pattern of visit is observed in the second case, that illustrates the capability of the E3D drive to set up specific polices devoted to the wide exploration of the environment.     

\begin{algorithm}[t]
	\caption{End-Effect Exploration Drive (E3D)}\label{algo:E3D}
	\begin{algorithmic}
		\REQUIRE{$\alpha$, $\beta$, $\lambda$, $\eta$}
		\STATE $Q \leftarrow \vec{0}_{|\mathcal{A}|\times n}$
		\STATE $p \leftarrow$ Uniform
		\STATE $p^* \leftarrow$ Uniform
		\WHILE{number of trials not exceeded}
			\STATE sample $a_{1:n} \sim \pi(A_{1:n})$ 
			\STATE read $s_n,r$
			\STATE $p \leftarrow (1-\eta) p + \eta \mathbf{1}_{S=s_n} $ 
			\FOR{$i \in 1..n$}
			    \STATE $Q(a_i) \leftarrow (1-\alpha\lambda) Q(a_i) + \alpha\lambda r - \frac{\alpha}{\beta} (\log p(s_n)-\log p^*(s_n))$ 
			\ENDFOR
		\ENDWHILE
	\end{algorithmic}
\end{algorithm}

\begin{figure}[t]
	\begin{subfigure}[b]{0.5\textwidth}
		\centerline{\bf (A)}
		\centerline{
			\includegraphics[width = \linewidth]{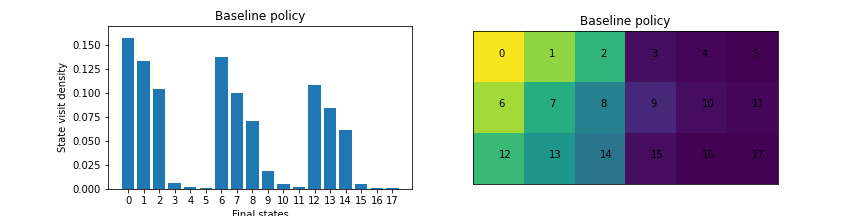} 		
		}
		\centerline{\bf (B)}
		\centerline{\includegraphics[width = \linewidth]{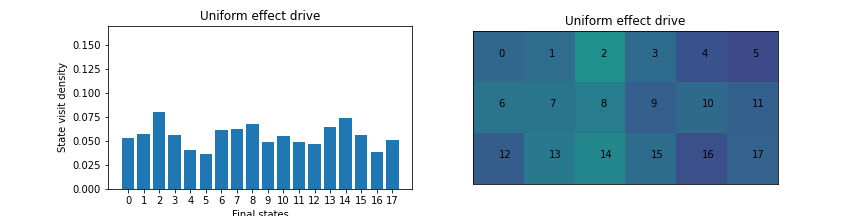} }
		%\label{fig:explore}
		\caption{Fig.~2. Task 1 : no reward is provided by the environment. Empirical distribution of the final states, after 5000 trials. {\bf A.} Uniform policy. {\bf B.} End-Effect Exploration Drive (E3D) algorithm. $\alpha=0.3$, $\beta = 1$, $\lambda=0.03$, $\eta=10^{-2}$.}
	\end{subfigure}
	\hspace{1cm}
	\begin{subfigure}[b]{0.4\textwidth}
		\centerline{\includegraphics[width=.8\textwidth]{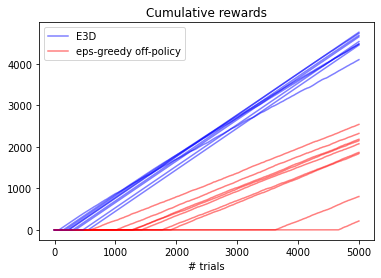}}
		%\label{fig:compare}
		\caption{Fig.~3. Task 2 : a reward $r=1$ is provided by the environment when the agent reaches the lower-right corner. Cumulative sum of rewards over 5000 trials, on 10 training sessions. The E3D algorithm is compared with state-of-the-art epsilon-greedy update. $\alpha=0.3$, $\beta = 100$, $\lambda=0.03$, $\eta=10^{-2}$, $\varepsilon=0.1$.}
	\end{subfigure}
\end{figure}

When a reward $r$ is provided by the environment, the question comes whether to balance the policy update procedure in favor of seeking for rewards or seeking for novel effects. By construction, the exploration drive is insensitive to the value of $\beta$, for the update is exactly proportional to $\frac{1}{\beta}$.  A high $\beta$ is associate with a small update and vice versa. this is not the case for the leftmost part of the update (equation \ref{eq:var-RL-lambda}). A high $\beta$ render the agent more sensitive to the extrinsic rewards. In practice, while no reward (or a uniform reward) is provided by the environment, the agent is only guided by the exploration drive. Once a reward is encountered, it tends to overtake the initial uniform exploration, providing a firm tendency toward a reward-effective selection of action. This is in contrast with the standard epsilon-greedy strategy, imposing to balance the exploration/exploitation trade-off by hand.

The E3D approach finally provides an Online/on-Policy training procedure that conforms to the main requirements of efficient reinforcement learning, showing both an efficient exploration policy when the rewards are sparse, and the capability to monitor the exploration/exploitation tradeoff with the inverse temperature $\beta$ in function of the magnitude of the rewards.

The cumulative rewards obtained with the E3D update and a state-of-the art off-policy/epsilon greedy update are compared in figure 3, %\ref{fig:compare},
with $\varepsilon$ set to 0.1. If both techniques manage to reach a reward-efficient policy in the long run, the exploration strategy developed in E3D makes it easier to reach the rewarding state, providing the reward earlier in time and developing a fast-paced reward-seeking strategy that overtakes the baseline approaches.

%\begin{figure}[t!]\label{fig:explore}
%	\centerline{\bf (A)}
%	\centerline{
%		\includegraphics[width = .6\linewidth]{../figures/botteneck-baseline.png} 		
%	}
%	\centerline{\bf (B)}
%	\centerline{\includegraphics[width = .6\linewidth]{../figures/botteneck-uniform-drive.png} }
%	\caption{Task 1 : no reward is provided by the environment. Empirical distribution of the final states, after 5000 trials. {\bf A.} Uniform policy. {\bf B.} End-Effect Exploration Drive (E3D) algorithm. $\alpha=0.3$, $\beta = 1$, $\lambda=0.03$, $\eta=10^{-2}$.}
%\end{figure}

%\begin{figure}[t!]\label{fig:compare}

%	\centerline{
%		\includegraphics[width = .4\linewidth]{../figures/training-comparison.png} 		
%	}
%	\caption{Task 2 : a reward $r=1$ is provided by the environment when the agent reaches the lower-right corner. Cumulative sum of rewards over 5000 trials, on 10 training sessions. The E3D algorithm is compared with state-of-the-art epsilon-greedy update. $\alpha=0.3$, $\beta = 100$, $\lambda=0.03$, $\eta=10^{-2}$, $\varepsilon=0.1$.}
%\end{figure}

\section{Conclusions}

Despite its simplicity, our pilot training setup allows to illustrate the main features expected from the inversion of a target distribution of effects, that is the capability to rapidly explore the environment through a reciprocal update of the policy and the effect model. We found in practice that the update of the model needs to be a bit slower than that of the policy to allow for the policy to improve over time and increase the extent of the effect space in a step by step manner. By balancing the effect of the rewards with the inverse temperature parameter, it is possible to catch and exploit very sparse rewards in large environments. 

The model is developed here as a first draft in an ``end-effect'' setup, with very little influence of the context or the states visited in the monitoring of the policy. Extensions toward closed-loop state-action policies is not far from reach, for equation (\ref{eq:var-RL-lambda}) allows to exploit the Bellman recurrence to guide the exploration-driven policy with a reference action-value function, that should be updated in parallel to the model and the current policy. Extensions toward continuous action spaces are also needed in order to address effective motor control learning, which resorts to a deeper interpretation of our approach toward the variational inference setup. 

%

\end{document}